# L'Apprentissage Automatique dans la planification et le contrôle de la production : un état de l'art

Juan Pablo Usuga Cadavid[1], Samir Lamouri[2], Bernard Grabot[3], Arnaud Fortin[4]

[1] LAMIH, Arts et Metiers ParisTech
151 Boulevard de l'Hôpital, 75013, Paris, France
juan_pablo.usuga_cadavid@ensam.eu

[2] LAMIH, Arts et Metiers ParisTech
151 Boulevard de l'Hôpital, 75013, Paris, France
samir.lamouri@ensam.eu

[3] LGP, ENIT
47 Avenue d'Azereix, 65000, Tarbes, France
bernard.grabot@enit.fr

[4] iFAKT France SAS
8 Esplanade compans caffarelli, 31000, Toulouse, France
a.fortin@ifakt.com

*Résumé* – La Planification et le Contrôle de la Production (PCP) sont essentiels pour assurer un avantage concurrentiel, réduire les coûts et respecter les dates de livraison. L'Apprentissage Automatique (AA) ouvre de nouvelles opportunités à la PCP pour prendre des décisions intelligentes à partir de données. Cette communication présente un état de l'art scientifique empirique sur l'utilisation de l'AA dans la PCP. Cette recherche a deux objectifs principaux : dans un premier temps, elle vise à identifier les techniques et outils utilisés pour implémenter l'AA dans la PCP, dans un second temps, elle analyse les caractéristiques de l'Industrie 4.0 (I4.0) fréquemment abordées dans la littérature scientifique. Concernant le second objectif, sept caractéristiques de l'I4.0 sont utilisées pour construire un cadre d'analyse, dont deux sont proposées par les auteurs. Les cas d'application de la PCP assistée par l'AA (PCP-AA) dans le cadre de l'I4.0 sont identifiés. Enfin, les résultats sont analysés et des perspectives de recherche sont proposées.

*Abstract -* Proper Production Planning and Control (PPC) is capital to have an edge over competitors, reduce costs and respect delivery dates. With regard to PPC, Machine Learning (ML) provides new opportunities to make intelligent decisions based on data. Therefore, this communication provides an initial systematic review of publications on ML applied in PPC. The research objective of this study is twofold: firstly, it aims to identify techniques and tools allowing to apply ML in PPC, and secondly, it reviews the characteristics of Industry 4.0 (I4.0) in recent research papers. Concerning the second objective, seven characteristics of I4.0 are used in the analysis framework, from which two of them are proposed by the authors. Additionally, the addressed domains of ML-aided PPC in scientific literature are identified. Finally, results are analyzed and gaps that may motivate further research are highlighted.

*Mots clés -* Apprentissage Automatique, Intelligence Artificielle, Industrie 4.0, Entreprise Intelligente, Planification et Contrôle de la Production

*Keywords -* Machine Learning, Artificial Intelligence, Industry 4.0, Smart Manufacturing, Production Planning and Control

## 1 Introduction

La génération de données dans les systèmes industriels modernes a connu une forte croissance jusqu'à atteindre autour de 1000 exaoctets par an [Tao et al., 2018]. Ces informations représentent une source précieuse de connaissances pour les industriels, car elles peuvent conduire à des améliorations et des économies. Néanmoins, le potentiel se trouvant dans ces données est insuffisamment exploité [Manns et al., 2015]. En conséquence, plusieurs pays ont proposé des plans visant l'adaptation de leurs industries aux nouveaux paradigmes technologiques. Notamment, l'Allemagne a introduit l'Industrie 4.0 (I4.0), les États-Unis ont créé la Coalition pour le leadership dans la fabrication intelligente et la Chine a proposé le plan Fabriqué en Chine 2025 [Wang et al., 2018a]. Ces initiatives se proposent de promouvoir la recherche sur l'industrie 4.0 à l'aide d'efforts financiers importants. En 2020, l'Union européenne aura investi autour de 7 milliards d'euros autour du concept des usines du futur [Kusiak, 2017].

Dans le cadre de l'I4.0, la PCP peut être définie comme la fonction qui définit les quantités globales à produire (plan de production) afin de respecter le plan commercial, tout en satisfaisant les objectifs généraux de rentabilité, de productivité, et de délais de livraison. Ceci inclut le contrôle du processus de production en permettant la synchronisation des ressources en temps réel ainsi que la personnalisation des produits [Tony Arnold et al., 2012; Kohler et Weisz, 2016].

Ruessmann et al. (2015) ont identifié neuf groupes de technologies permettant l'implémentation de l'I4.0. Parmi ces neuf éléments, cette recherche se concentre sur le *Big Data Analytics* (BDA), et plus spécifiquement sur l'AA appliqué en PCP. La définition de l'AA qui est retenue est celle proposée par Mitchell (1997), qui fait référence à un programme d'ordinateur capable d'apprendre de l'expérience afin d'améliorer une mesure de performance dans l'exécution d'une tâche donnée.

Même s'il existe des méthodes analytiques efficientes pour réaliser la PCP, les solutions proposées deviennent rapidement infaisables dans la phase d'exécution. Cela est causé par l'incertitude (pannes de machine, taux de rebut, etc.) et la nature stochastique des processus manufacturiers. De plus, les Progiciels de Gestion Intégrée (PGI) ont une mauvaise performance au niveau opérationnel [Gyulai et al., 2015]. Enfin, les marchés volatils actuels, caractérisés par une forte tendance à la personnalisation de masse et à un respect strict des dates de livraison, imposent une complexité nécessitant une PCP robuste [Reuter et al., 2016]. L'AA peut améliorer la robustesse de la PCP, car les connaissances se trouvant dans les données peuvent aider à gérer les évènements prévisibles et imprévisibles.

Ayant reconnu la contribution potentielle de l'AA pour la PCP, le but de cette communication est de mener un état de l'art systématique de la littérature scientifique sur l'utilisation de l'AA dans la PCP, dans le cadre de l'I4.0. Trois questions de recherche sont abordées :

1. Quels sont les cas d'application de la PCP-AA traités par la littérature scientifique récente ?
2. Quels sont les techniques et les outils utilisés pour implémenter une PCP-AA ?
3. Quelles sont les caractéristiques de l'I4.0 fréquemment abordées par les articles scientifiques ?

La première question vise à identifier les cas d'application de la PCP-AA les plus étudiés par la recherche empirique. Cela permettra de repérer des tendances et des perspectives de recherche. La deuxième et la troisième question répondent aux deux objectifs de recherche proposés par cette étude. La deuxième question cherchera à identifier les techniques et les outils : ces éléments seront utiles pour construire, dans des recherches futures, une méthode pour appliquer l'AA dans la PCP. Enfin, la troisième question permettra de comprendre quelles sont les caractéristiques de l'I4.0 les plus et les moins abordées, ce qui nous permettra de proposer des perspectives de recherche.

La section 2 présentera la méthode utilisée pour mener un état de l'art systématique. La section 3 expliquera le cadre d'analyse qui sera utilisé. Ensuite, la section 4 portera sur les résultats de la revue de littérature systématique et leur analyse. La section 5 conclura ce travail et proposera des perspectives de recherche.

## 2 METHODOLOGIE DE RECHERCHE

Afin de répondre aux trois questions proposées, une étude de la littérature scientifique a été menée. Un état de l'art systématique de la littérature suivant la méthodologie de recherche proposée par Tranfield et al. (2003) a été réalisée. Cet état de l'art est focalisé exclusivement sur des articles d'application empirique de l'AA dans la PCP dans le cadre de l'I4.0.

Les bases de données analysées ont été ScienceDirect et SCOPUS. Les requêtes ont été réalisées entre le 10/10/2018 et le 05/11/2018. Pour s'assurer de rester dans le contexte de l'I4.0, seuls les articles publiés à partir de 2011 ont été pris en compte. Cette année correspond en effet à l'introduction du terme « I4.0 » à la foire d'Hanovre. Les mots clefs suivants ont conduit la recherche dans les titres (*titles*), résumés (*abstract*) et mots clefs (*keywords*) :

- (« Deep Learning » OR « Machine Learning ») AND (« Production scheduling »)
- (« Deep Learning » OR « Machine Learning ») AND (« Production control »)
- (« Deep Learning » OR « Machine Learning ») AND (« Line balancing »)
- (« Deep Learning » OR « Machine Learning ») AND (« Production planning »)

En plus de la restriction de l'année de publication (année >= 2011), seulement des *« Articles de recherche »* sur ScienceDirect et des *« Articles de conférence »* ou des *« Articles »* sur SCOPUS ont été retenus. Ensuite, une révision des titres et des résumés a permis une première sélection des applications empiriques d'AA en PCP. Les articles doublons ont été enlevés. Finalement, une analyse de tout le texte pour chaque article retenu a permis une seconde sélection et la constitution de l'échantillon final pour mener la recherche. La figure 1 synthétise cette méthodologie de recherche.

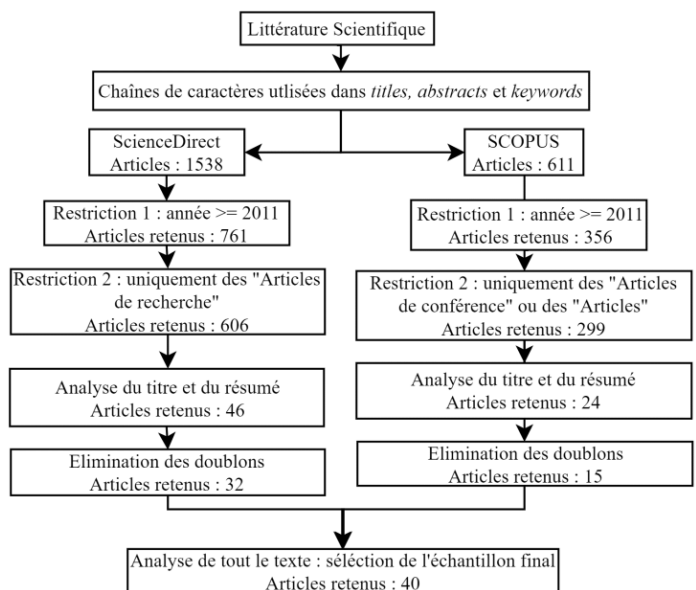

**Figure 1. Méthodologie de recherche utilisée**

Les 40 articles sélectionnés ont été étudiés suivant le cadre d'analyse proposé dans la section suivante.

## 3 CONSTRUCTION DU CADRE D'ANALYSE

Le cadre d'analyse proposé est constitué de trois composants. Chaque composant vise à répondre à une des questions de recherche mentionnées auparavant.

## 3.1 Premier composant du cadre d'analyse : les cas d'application de la PCP-AA dans l'I4.0

Identifier des cas d'application et la façon dont ils sont considérés dans la littérature scientifique est important pour mieux identifier les perspectives de recherche. Par cas d'application, cette étude fait référence à des usages de la PCP-AA sur un domaine industriel, comme l'ordonnancement, le contrôle de qualité, la maintenance, etc. Tao et al. (2018) ont identifié six cas d'application du BDA dans l'I4.0 :

1. La Maintenance Intelligente (MI) : exploitation de données pour faire de la maintenance préventive ou prédictive. Par exemple, estimer le moment de réaliser une intervention de maintenance en surveillant l'état des composants de la machine. Dans le cas de la maintenance prédictive, l'analyse peut même mener à la découverte de la cause racine du défaut du composant.

2. Le Contrôle de Qualité des Produits (CQP) : application du BDA pour surveiller la fabrication, à la recherche de la non-qualité. En plus de l'identification de défauts, cela concerne également la détection des non-conformités dès leur moment d'apparition et la détermination de leur cause racine.

3. Le Monitoring du Processus de Production (MPP) : utilisation des données pour réaliser un paramétrage intelligent des ressources physiques (machines, chariots, entre autres) du processus de production. Le but est de faire fonctionner les ressources automatiquement et/ou de manière optimisée par rapport aux conditions de travail.

4. La Distribution et le Suivi de Pièces (DSP) : gestion de stocks, géolocalisation des pièces et gestion de la distribution de manière intelligente à l'aide des données historiques et en temps réel. Par exemple, lancer automatiquement et suffisamment en avance un ordre de réapprovisionnement quand un système estime que les pièces détachées d'un stock lié à un poste de travail vont s'épuiser.

5. La Planification et l'Ordonnancement Intelligents (POI) : prise en compte des aléas de production pour réaliser une planification et un ordonnancement plus proches de l'état actuel du système. Par exemple, prendre en compte les pannes machine ainsi que les problèmes de qualité pour réaliser l'ordonnancement d'une commande client.

6. La Conception Intelligente (CI) : usage du BDA pour supporter la conception. Cela comprend des activités allant de la conception d'un nouveau produit jusqu'à la conception d'une gamme de fabrication.

L'analyse des 40 articles a montré que ces 6 éléments n'étaient pas suffisants pour classer certains articles. De plus, ces articles partageaient un même domaine d'application : l'estimation de temps (temps de cycle, temps d'opération, etc.). De ce fait, un cas d'application supplémentaire est proposé :

7. L'Estimation de Temps (ET) : adaptation des temps opératoires, temps de cycle, temps d'attente et autres, aux conditions réelles de travail. Par exemple, adapter le temps opératoire à la cadence réelle de chaque employé au lieu d'utiliser les valeurs moyennes d'une étude de temps.

## 3.2 Deuxième composant du cadre d'analyse : les éléments d'une méthode

La deuxième question de cette recherche est liée aux techniques et outils pour implémenter une PCP-AA. Selon Zellner (2011), ces deux éléments sont liés au concept d'« Eléments Indispensables d'une Méthode ». Ainsi, il a identifié cinq éléments fondamentaux qui constituent une méthode :

1. Une procédure : suite d'activités à réaliser au moment d'appliquer la méthode.

2. Des techniques : le moyen pour générer des résultats. Les activités sont supportées par les techniques et ces dernières sont supportées par les outils.

3. Des résultats : ce qui est créé ou fourni par une activité.

4. Des rôles : le point de vue adopté par la personne qui réalise une activité et qui en est responsable.

5. Un modèle d'information : relation entre les quatre éléments mentionnés auparavant.

Dans le cadre de cette recherche, le deuxième élément (les techniques) est concerné. Le but est d'identifier les techniques et les outils pour les mettre en place au moment d'implémenter une PCP-AA. Plus spécifiquement, ces techniques font référence à des modèles d'AA, par exemple les Réseaux de Neurones (RN) ou les Forêts d'Arbres décisionnels (FA). Les outils feront référence aux langages ou logiciels utilisés pour implémenter un modèle d'AA : Python, R, RapidMiner, etc.

Pour complémenter l'analyse des techniques, les types d'apprentissages seront aussi évalués. Cela permettra de synthétiser l'information et d'identifier plus facilement des perspectives de recherche ainsi que des tendances. En s'appuyant sur les travaux de Jordan et Mitchell (2015), les types d'apprentissages peuvent être définis de la manière suivante :

1. Apprentissage Supervisé (AS) : ce sont des techniques qui font une approximation d'une fonction $f(X)$ à partir de l'apprentissage de la relation entre les données d'entrée $X$ et de sortie $Y$. Par exemple, apprendre la relation entre les pixels d'une image (entrée $X$) et l'objet auquel cela correspond (sortie $Y$), pour déterminer si une nouvelle image correspond à celle d'un vélo.

2. Apprentissage Non supervisé (AN) : il correspond aux techniques permettant l'exploration et la découverte de structures dans un ensemble de données $X$. Par exemple, identifier des segments de marché dans un ensemble de données $X$ contenant l'âge et le montant des dépenses des clients dans un magasin.

3. Apprentissage par Renforcement (AR) : il s'agit des techniques visant l'apprentissage d'actions par un agent qui interagit avec un environnement pour maximiser une récompense. Par exemple, apprendre à un robot à marcher en récompensant les actions qui maximisent la distance horizontale parcourue.

*3.3 Troisième composant du cadre d'analyse : les caractéristiques de l'I4.0*

L'I4.0 vise à transformer les données acquises tout au long du cycle de vie du produit en « intelligence » pour améliorer le processus manufacturier [Tao et al., 2018]. Ces améliorations concernent la qualité, la productivité et la durabilité du système de production, tout en cherchant une réduction de coûts [Wang et al., 2018a]. Il est capital d'identifier les caractéristiques de l'I4.0 afin d'évaluer qualitativement sa maturité au sein d'une société. Tao et al. (2018) mentionnent que l'I4.0 est caractérisée par sa faculté à faciliter les paradigmes suivants :

1. Les produits centrés sur le client (cCl) : le système de production s'ajuste en prenant en compte des variables liées aux clients, comme leur comportement, leurs besoins, la façon d'utiliser les produits, entre autres. Des exemples d'application sont la fabrication de produits personnalisés, la génération de gammes de production à partir des commandes client, la prise en compte du comportement de chaque client pour proposer un prix de vente attractif, entre autres.

2. L'auto-organisation des ressources et des tâches (aOr) : les données provenant du processus manufacturier sont utilisées pour réaliser une planification de la production et une allocation de ressources plus intelligentes. Un exemple serait l'ordonnancement de tâches et l'équilibrage de charges de production de façon dynamique. Autrement dit, l'ordonnancement et l'équilibrage prennent en compte l'avancement de la production en temps réel pour répondre aux aléas.

3. L'auto-exécution des ressources et des processus (aEx) : les ressources (machines, ordinateurs, chariots de livraison, etc.) opérant dans la production sont capables de prendre des décisions à partir des informations mesurées ou reçues. C'est le cas par exemple de machines qui adaptent leurs paramètres de fonctionnement de façon automatique ou de chariots qui réapprovisionnent des pièces de manière autonome.

4. L'auto-régulation du processus de production (aRe) : le système de production est capable de s'adapter aux évènements inattendus. Par exemple, l'arrivée d'une nouvelle commande client ou la panne soudaine d'une machine relancent automatiquement le processus de planification et d'ordonnancement de la production.

5. L'auto-apprentissage et l'adaptation du processus de production (aAp) : le système de production analyse en continu des variables externes et internes afin de prédire des aléas. C'est le cas de la maintenance prédictive, qui utilise des techniques de BDA pour calculer la durée de vie des composants des machines. De ce fait, le système est capable d'apprendre des aléas passés pour prédire l'apparition d'un problème.

Après l'analyse de l'échantillon, malgré l'importance des concepts abordés par ces cinq caractéristiques, deux éléments nous paraissent avoir été laissés de côté : l'aspect environnemental et la génération de connaissances. Pour en tenir compte, deux caractéristiques supplémentaires sont proposées :

6. Les processus de production centrés sur l'environnement (cEn) : les techniques de BDA sont utilisées à des fins environnementales en entreprise. Cela peut aller de l'ajustement des paramètres machine pour diminuer la consommation d'énergie jusqu'à l'ordonnancement d'un processus de recyclage afin de maximiser le nombre de composants récupérés dans le minimum de temps.

7. La découverte et la génération de connaissances (gKn) : l'exploitation de données permet d'améliorer les connaissances de type métier et de créer des nouvelles notions sur le fonctionnement du système de production. Par exemple, l'extraction de Règles d'Association des bases de données de maintenance peut permettre d'identifier des corrélations aidant l'entreprise à s'améliorer. Dans cette caractéristique, l'interprétabilité des résultats est un facteur clef.

La section suivante présente le classement des différents articles selon les trois composants du cadre d'analyse.

## 4 RESULTATS

*4.1 Première question de recherche : les cas d'application*

Pour répondre à cette question, chaque article analysé a été relié à un cas d'application. Cela a permis de mesurer son utilisation (figure 2).

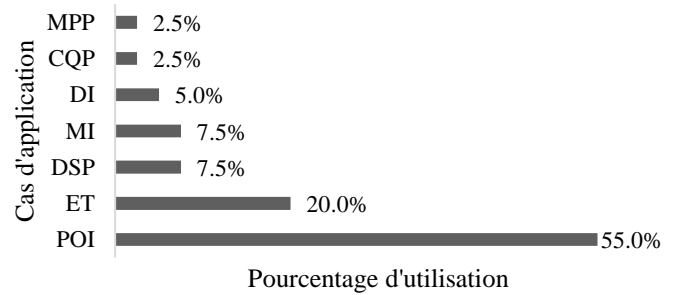

**Figure 2. Utilisation des cas d'application de la PCP-AA**

Les résultats montrent que les cas d'application les plus souvent liés à la PCP-AA sont le POI et l'ET. Cela était attendu, car la planification et l'ordonnancement de la production ainsi que l'estimation de temps sont des domaines classiquement liés à la PCP. Néanmoins, il semble y avoir un grand manque de recherche dans les cinq autres cas d'application. Notamment, le MPP et le CQP sont les deux cas les moins considérés, avec 2,5 % des articles.

Dans les dernières années, des cas d'application comme le CQP et la MI ont reçu beaucoup d'attention pour l'utilisation de l'AA. Malgré cela, la liaison entre l'AA, l'un de ces deux cas d'application et la PCP est rare dans la littérature scientifique. En outre, les domaines liés à la conception des processus (CI) et à la géolocalisation d'outils de production (DSP) commencent à être abordés par l'AA et liés à la PCP. Cependant, les efforts de recherche doivent être accrus du fait de leur faible utilisation.

*4.2 Deuxième question de recherche : les techniques et les outils*

Concernant les techniques, le nombre de fois que chaque technique a été utilisée est mesuré. Dans le cas des articles utilisant plusieurs techniques, seulement celle considérée par l'auteur ou les auteurs comme la plus performante est retenue. Pour les articles utilisant un modèle qui a recours à plusieurs techniques, toutes les techniques concernées sont retenues. Par exemple, Rostami et al. (2018) proposent un modèle employant les Machines à Vecteurs de Support (MVS), les k-Moyennes (k-Means) et l'Analyse de Composantes Principales (PCA). Dans ce cas, chaque technique est comptabilisée comme utilisée une fois.

En raison du grand nombre de techniques trouvées, un regroupement par famille est proposé afin de permettre une meilleure analyse des résultats. Ces familles de techniques ont été construites avec l'aide et la validation d'un expert du domaine. Le tableau 1 présente l'ensemble de techniques incluses dans chaque famille. Il faut mentionner que la colonne « techniques incluses » contient uniquement celles trouvées dans l'état de l'art ; la liste n'est donc pas exhaustive.

**Tableau 1. Familles et leurs techniques**

| Famille | Techniques incluses |
|---|---|
| Règle d'Association | Règle d'Association |
| Modèle Bayésien | Classification bayésienne naïve, Réseau Bayésien |
| Clustering | K-Moyennes, Regroupement flou, Regroupement par des pics de densité, Regroupement hiérarchique |
| k-NN | K plus proches voisins |
| Réseaux de Neurones (RN) | Perceptron multicouche à rétropropagation, auto-encodeur, machine d'apprentissage extrême, Long Short-Term Memory (LSTM) |
| Analyse en Composantes Principales (PCA) | Analyse en Composantes Principales |
| Q-Learning | Q-Learning |
| Régression | Régression polynomiale, Régression linéaire, Régression par des fonctions de base radiale |
| Sarsa | Sarsa |
| Machines à Vecteurs de Support (MVS) | Machines à Vecteurs de Support |
| Modèles à base d'Arbres de Régression et Classification (CART) | Arbres de décision, Forêts aléatoires (FA) |

Les résultats de l'utilisation de chaque famille sont présentés dans la figure 3. Ils indiquent que les CART et les RN restent les familles de modèles les plus utilisées dans l'implémentation d'une PCP-AA. En effet, les premiers ont un bon compromis entre l'exactitude et l'interprétabilité du modèle, tandis que les seconds sont très performants en présence de données présentant des relations non linéaires.

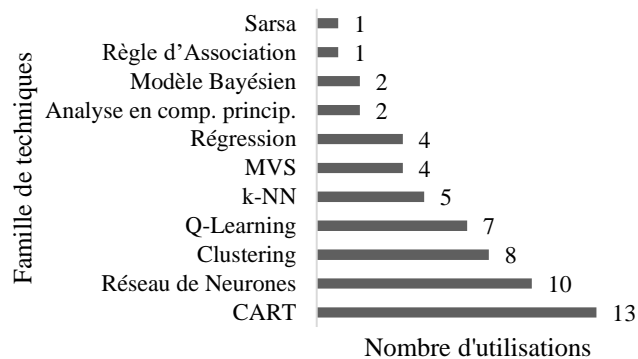

**Figure 3. Nombre d'utilisations par famille de techniques**

L'expérience montre que les techniques de Clustering sont aussi très utilisées. Cela est probablement dû au fait qu'elles permettent de gérer les données non étiquetées, abondantes dans le contexte manufacturier. En outre, des techniques d'apprentissage par renforcement comme Sarsa ou Q-Learning ont été largement employées, ce qui dénote un intérêt pour les modèles multi-agents en PCP.

Il est étonnant de constater que la PCA et les Règles d'Association sont rarement appliquées dans la PCP-AA (leur usage se limite à deux et une fois, respectivement). Ce résultat est surprenant, car ces deux techniques présentent de forts avantages. En effet, la PCA permet de réaliser, entre autres, un prétraitement de données et les Règles d'Association permettent d'extraire des bases de données des connaissances précieuses.

Les types d'apprentissages ont été comptabilisés. Étant donné qu'un article peut utiliser plusieurs techniques d'AA, il peut faire référence à plusieurs types d'apprentissages en même temps. Les combinaisons possibles entre les différents types d'apprentissages ont été prises en compte. La figure 4 présente ces résultats :

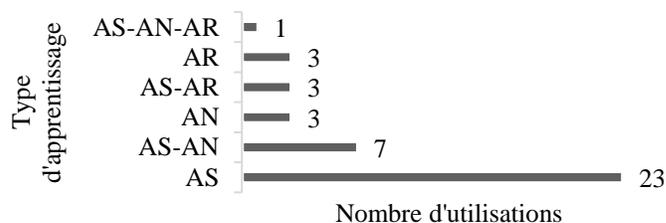

**Figure 4. Nombre d'utilisations par type d'apprentissage**

L'expérience permet de constater que le type d'apprentissage le plus utilisé est l'AS. Cela est probablement dû au fait qu'il permet de satisfaire des besoins récurrents dans la recherche et l'industrie : la classification et la régression. En effet, les techniques d'AS peuvent apprendre des relations entre des entrées $X$ et des sorties $Y$ continues, dans le cas de la régression, ou des sorties $Y$ discrètes, dans le cas de la classification. Ainsi, il suffit d'avoir besoin de déterminer des catégories ou d'approximer une fonction inconnue pour faire appel à l'AS, d'où sa forte utilisation.

Les résultats montrent aussi que l'AS et l'AN (AS-AN) sont souvent couplés. Ceci peut être dû au fait que l'AN fournit des outils qui permettent le prétraitement de données, comme la PCA, ainsi que l'identification de catégories dans des bases de données avec les techniques de Clustering. Par ailleurs, on peut constater que l'utilisation pure d'AN et d'AR ainsi que d'AN-AR est faible. Finalement, l'usage des trois types d'apprentissages en même temps est très rare, avec un seul article dans l'échantillon.

Concernant les outils, seulement les logiciels utilisés pour implémenter le modèle d'AA sont pris en compte. De ce fait, les logiciels de simulation à évènements discrets ne sont pas inclus dans cette recherche. Les résultats sont présentés dans la figure 5.

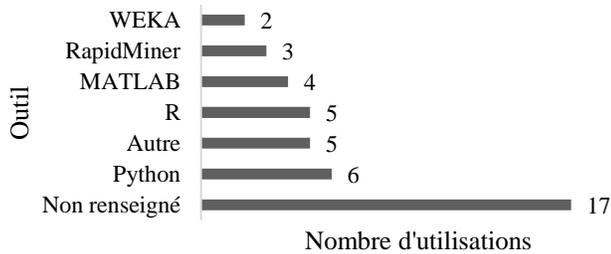

**Figure 5. Nombre d'utilisations par outil**

Pour des raisons de clarté d'affichage, les outils utilisés une seule fois ont été inclus dans une catégorie « Autre ». Ces outils sont : ACE Datamining System, Clementine, Neural-SIM, Visual C++ et Xelopes Library. La plupart des articles scientifiques ne mentionnent pas l'outil qu'ils utilisent pour créer leurs modèles.

Les autres résultats indiquent une forte tendance à l'utilisation de trois outils principaux : Python, R et MATLAB, avec six, cinq et quatre utilisations respectivement. Par ailleurs, il est important de mentionner que dans les six usages de Python, trois sont implémentés avec des librairies spécialisées en AA et Apprentissage Profond (AP). Ces librairies sont TensorFlow (deux usages) et Keras (un usage). Effectivement, l'usage des librairies spécialisées est important, car il permet des traitements plus rapides et optimisés. De plus, l'implémentation des techniques plus complexes d'AP comme les LSTM est facilitée.

*4.3 Troisième question de recherche : les caractéristiques de l'I4.0*

Pour répondre à la troisième question, le modèle de PCP-AA proposé par chaque article été analysé et les caractéristiques de l'I4.0 (présentées dans le cadre d'analyse) qu'il concernait ont été comptabilisées. De ce fait, un article donné peut impliquer les sept caractéristiques en même temps. Le tableau 2 montre les résultats pour chaque article de l'échantillon, ainsi que le nombre de fois que chaque caractéristique est abordée. La première colonne à gauche du tableau 2, appelée « N. » (numéro), assigne un numéro à chaque article analysé. Cela sera utilisé dans la figure 6 pour des questions de simplification et clarté du graphe.

**Tableau 2. Caractéristiques de l'I4.0 par article**

| N. | Reference | cCl | aOr | aEx | aRe | aAp | cEn | gKn |
|---|---|---|---|---|---|---|---|---|
| 1 | [Altaf et al., 2018] | | X | | | | | X |
| 2 | [Diaz-Rozo et al., 2017] | | | | | X | | X |
| 3 | [Dolgui et al., 2018] | | | | | X | | X |
| 4 | [Gyulai et al., 2014] | | X | | | | | X |
| 5 | [Gyulai et al., 2015] | | X | | X | | | |
| 6 | [Gyulai et al., 2018] | | X | | | X | | |
| 7 | [Hammami et al., 2016] | | | | X | X | | |
| 8 | [Ji et Wang, 2017] | | X | | X | | | |
| 9 | [Kartal et al., 2016] | | | | | | | X |
| 10 | [Kho et al., 2018] | | | | | | | X |
| 11 | [Kruger et al., 2011] | | | X | | | | X |
| 12 | [Lai et Liu, 2012] | | X | | | | | X |
| 13 | [Leng et al., 2018] | X | X | | | | | |
| 14 | [Li et al., 2012b] | X | X | | | | | |
| 15 | [Li et al., 2012a] | | | | | X | | X |
| 16 | [Lingitz et al., 2018] | | X | | | | | |
| 17 | [Lubosch et al., 2018] | | X | | | | | |
| 18 | [Lv et al., 2018] | | X | | | | | |
| 19 | [Manns et al., 2015] | | X | | | | | X |
| 20 | [Mori et Mahalec, 2015] | | X | | | | | |
| 21 | [Palombarini et Martínez, 2012] | | X | | X | | | |
| 22 | [Qu et al., 2016] | | X | | | | | |
| 23 | [Reboiro-Jato et al., 2011] | | X | | | | | |
| 24 | [Reuter et al., 2016] | | | | X | | | |
| 25 | [Rostami et al., 2018] | | | | X | | | X |
| 26 | [Schuh et al., 2017] | | X | | | X | | X |
| 27 | [Shahzad et Mebarki, 2012] | | | | | | | X |
| 28 | [Solti et al., 2018] | | | | X | | | |
| 29 | [Stricker et al., 2018] | | X | | X | X | | |
| 30 | [Tian et al., 2013] | | X | | | | X | |
| 31 | [Tong et al., 2016] | | X | | | | X | |
| 32 | [Tuncel et al., 2012] | | X | | | | X | |
| 33 | [Wang et al., 2015] | | X | | | | | |
| 34 | [Wang et al., 2018c] | | | | X | | | |
| 35 | [Wang et al., 2018b] | | | | X | | | |
| 36 | [Waschneck et al., 2018] | | X | | | | | |
| 37 | [Wauters et al., 2012] | | | | | X | | X |
| 38 | [Yang et al., 2016] | | X | | X | | | |
| 39 | [Zhang et al., 2011] | | X | | | | | |
| 40 | [Zhong et al., 2013] | | X | | | | | |
| | Totaux | 2 | 26 | 1 | 10 | 9 | 3 | 14 |

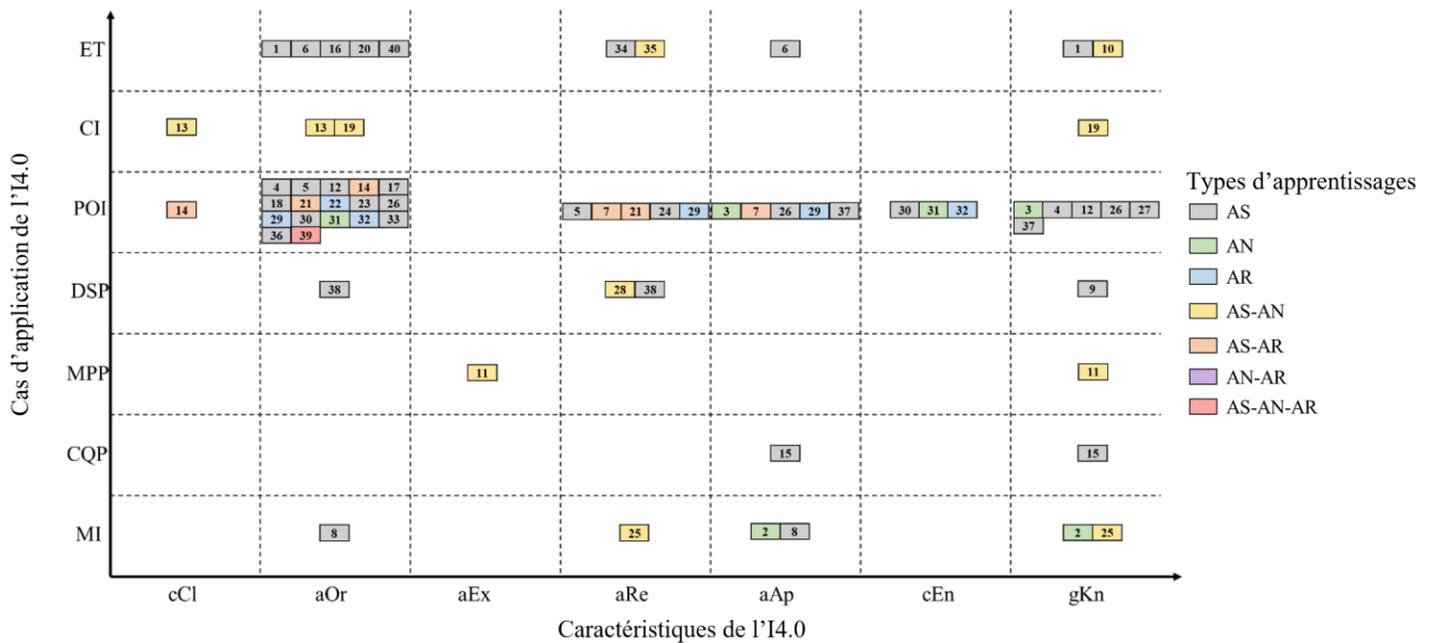

**Figure 6. Matrice croisée entre les cas d'application, les caractéristiques et les types d'apprentissages**

Les résultats montrent que l'aOr est, de loin, la caractéristique la plus considérée (26 usages) par les articles empiriques portant sur la PCP-AA. Cela était attendu puisque l'aOr concerne la planification de la production et l'ordonnancement intelligents, sujets directement liés à la PCP. Par ailleurs, cela indique que l'AA est capable de satisfaire cette caractéristique au sein de la PCP.

Trois caractéristiques peuvent être considérées comme moyennement satisfaites : la gKn, l'aRe et l'aAp. Notamment, la gKn se présente comme la deuxième caractéristique la plus retrouvée (14 usages). Cela suggère qu'une grande partie de la recherche empirique dans le domaine vise à extraire des connaissances au moyen de l'AA. En conséquence, il est possible d'affirmer que l'AA s'avère utile pour faire cette extraction, ce qui est fondamental dans l'I4.0, dans laquelle les données sont abondantes et peuvent fournir des connaissances métier précieuses.

L'aRe et l'aAp sont liées à la capacité du système à s'adapter aux évènements imprévus et prévus, respectivement. Leur utilisation, 10 usages pour l'aRe et 9 pour l'aAp, permet de constater deux choses : premièrement, l'AA propose un moyen de prendre en considération l'environnement dynamique manufacturier dans lequel il existe des aléas inattendus. Deuxièmement, la PCP commence à bénéficier de la capacité des techniques d'AA à prédire l'occurrence des évènements à partir des données.

Malgré son utilité pour la PCP-AA, trois caractéristiques n'ont pas été suffisamment abordées : 35 % des articles pour la gKN, 25 % pour l'aRe et 22,5 % pour l'aAp. Cette faible utilisation suggère des perspectives de recherche pour atteindre plus de maturité dans l'I4.0 avec la PCP.

Finalement, trois caractéristiques ont été très rarement utilisées : les cCl (2 usages), l'aEx (1 usage) et les cEn (3 usages). Trois conclusions peuvent être proposées : premièrement, cela montre que très peu de recherches font la liaison entre la PCP, l'AA et l'analyse des variables client. Dans un second temps, la très faible utilisation de l'aEx montre un manque important dans le développement de machines et de robots autonomes liés avec la PCP-AA. En effet, la littérature scientifique en robotique est vaste, mais peu d'applications ont cherché à utiliser l'AA en robotique pour améliorer la PCP. Enfin, le fait que les cEn ont été abordés seulement trois fois induit de grandes perspectives de recherche. Parmi ces trois usages, deux visent l'ordonnancement de processus de désassemblage pour améliorer le recyclage des pièces. L'article restant traite de l'optimisation énergétique dans une forge.

*4.4 Analyse croisée entre les trois composants du cadre d'analyse*

Afin de synthétiser les résultats et de fournir un aperçu global du cadre d'analyse, la figure 6 est proposée. Elle montre les articles de l'échantillon placés dans un espace à deux dimensions : les caractéristiques de l'I4.0 (sur l'axe horizontal) et les cas d'application de l'I4.0 (sur l'axe vertical). Les articles ont été numérotés dans le tableau 2, ce qui a permis de les situer sur le schéma. Concernant le composant lié aux techniques, les types d'apprentissages sont représentés sur le graphique au moyen d'un code couleur.

Ce schéma a deux buts principaux : d'une part, il cherche à montrer quelles sont les caractéristiques les plus considérées, dans quels cas d'application et avec quel type d'apprentissage automatique. D'autre part, il permet de visualiser les perspectives de recherche d'un coup d'œil.

Tout d'abord, si chaque case (une caractéristique dans un cas d'application donné) représente un domaine potentiel de la PCP-AA, il est possible de dire qu'il existe 49 (7 cas d'application x 7 caractéristiques) domaines potentiels qui peuvent être considérés dans l'I4.0. Si les cases ou domaines ayant au moins un article attribué sont considérés comme « couverts », il est possible d'affirmer qu'il n'y a que 24 domaines couverts par la littérature scientifique actuelle. Cela représente un taux de couverture de

49 %, montrant un fort besoin de recherche dans la PCP-AA en général.

Les résultats permettent de constater que les articles appartenant au cas d'application le plus utilisé, le POI, répondent à presque toutes les caractéristiques de l'I4.0, sauf à l'aEx et très peu aux cEn ainsi qu'aux cCl. En effet, lier des machines autonomes, des variables environnementales ou des variables provenant du client à la planification et l'ordonnancement de la production reste rare dans la littérature. Cette tendance est la même dans les six autres cas d'application. De cette manière, les résultats signaleraient des perspectives de recherche dans l'application de PCP-AA dans l'aEx, les cEn et les cCl.

L'AR et ses usages en synergie avec d'autres types d'apprentissages ont été utilisés uniquement dans le POI. Effectivement, il s'agit des techniques liées à la modélisation multi-agents, ce qui s'avère utile au moment de modéliser un environnement complet de production. Cependant, l'AR pourrait fournir un moyen pour davantage aborder l'aEx, car il permet le développement de machines autonomes. Malgré cela, les applications restent complexes à implémenter.

L'expérience montre une utilisation répandue de l'AS dans tous les cas d'application et caractéristiques de l'I4.0, qu'il soit employé seul ou avec un autre type d'apprentissage. Par ailleurs, l'AS est souvent accompagné par l'AN (AS-AN : 7 usages), ce qui pourrait signaler une synergie potentiellement intéressante. Cela contraste avec une faible utilisation de l'AN pur (3 usages). En effet, l'AN se présente le plus souvent comme un moyen pour prétraiter les données ou les explorer, d'où son usage avec d'autres types d'apprentissages.

Finalement, la seule caractéristique qui a été abordée par tous les cas d'application a été la gKn. Cela indique un intérêt général pour la génération de connaissances métier à partir des données. Par ailleurs, cette caractéristique montre une forte présence d'AN, d'AS et de leur synergie, signalant une forte affinité entre ces types d'apprentissages et la découverte d'information exploitable.

## 5 CONCLUSION

Cet état de l'art a été développé autour de trois axes principaux : les cas d'application de l'I4.0, les techniques et les outils pour réaliser une PCP-AA, et finalement, les caractéristiques de l'I4.0. A partir de cela, une analyse a été menée au moyen de l'examen d'un échantillon de 40 articles empiriques appliquant la PCP-AA. Les articles de l'échantillon ont été choisis en suivant la méthodologie d'une révision systématique de littérature proposée par Tranfield et al. (2003). En plus de l'identification de tendances et de perspectives de recherche, cette étude présente aussi un apport à la littérature avec l'identification d'un nouveau cas d'application et de deux nouvelles caractéristiques de l'I4.0.

Dans le cadre de la PCP-AA, les premiers résultats ont montré que les cas d'application les plus abordés par la littérature récente sont le POI et l'ET, tandis que les cinq autres cas restent faiblement traités. Effectivement, le POI et l'ET sont des cas d'application directement concernés par la PCP, d'où probablement son utilisation élevée. Cependant, le fait que les autres cas d'application restent rarement concernés exprime un besoin de diversification des autres applications de la PCP-AA. Cela permettrait de couvrir davantage de besoins industriels à partir de la PCP.

Les familles de techniques les plus utilisées sont les CART, les RN et le Clustering. La raison de leur forte utilisation est peut-être liée au fait que chacune présente des précieux avantages : les CART sont relativement simples à implémenter, avec un bon niveau d'interprétabilité et d'exactitude, ce qui s'avère utile pour la gKn. Les RN ont été fréquemment utilisées dans les dernières années à la suite de l'arrivée de l'AP. Les RN permettent de traiter des données qui présentent des structures non linéaires tout en gardant une bonne exactitude des résultats. Néanmoins, l'interprétabilité du modèle est négativement affectée à cause de son architecture complexe. De cette manière, on peut dire que les RN sont des techniques utilisées quand l'application est axée sur l'exactitude plutôt que sur l'interprétabilité. Finalement, les techniques de Clustering ont trouvé une place privilégiée au sein de la PCP-AA car elles permettent, entre autres, de découvrir des structures cachées dans les données. Cela est de grande utilité pour étiqueter des données brutes et entraîner ensuite des techniques d'AS, besoin très récurrent dans l'application industrielle de l'AA.

Les techniques d'AR comme Q-Learning et Sarsa commencent à s'utiliser, ce qui montre un intérêt pour les systèmes multi-agents dans la PCP-AA. Néanmoins, la figure 6 permet de constater que leurs applications se trouvent uniquement dans le POI et aucune d'entre elles n'a ciblé l'aEx ou la gKn. En effet, l'AR pourrait contribuer au développement de machines autonomes, mais ces applications restent complexes et chères à implémenter. Par ailleurs, le fait qu'aucune application d'AR ne cible la caractéristique de la gKn suggère un mauvais niveau d'interprétabilité. En conséquence, une perspective de recherche très porteuse serait le développement de modèles d'AR permettant une meilleure interprétation de leurs résultats.

Le type d'apprentissage utilisé par la plupart des articles est l'AS. Cependant, des synergies entre les types d'apprentissages commencent à apparaître. Notamment, la synergie entre l'AS et l'AN a été souvent utilisée. Elle permet d'explorer et de prétraiter les données au moyen de l'AN, pour ensuite représenter des relations mathématiques complexes avec l'AS. Aucun article ne s'est servi de la synergie entre l'AN et l'AR, ce qui suggère un besoin d'intégration entre ces deux types d'apprentissages. En outre, seulement un article a utilisé les trois types en même temps.

Chaque type d'apprentissage présente des avantages et des inconvénients propres. De ce fait, créer des synergies entre eux permettrait probablement de repousser leurs limites. Ainsi, il est important de faire de la recherche dans la PCP-AA intégrant les différents types d'apprentissages.

L'utilisation des outils dans l'échantillon suggère une prévalence de langages de programmation comme Python et R. De plus, l'usage de Python a montré que les articles récents commencent à employer des libraries spécialisées comme TensorFlow et Keras pour faciliter la création de modèles plus complexes. En outre, les chercheurs se servent de plus en plus de logiciels comme RapidMiner permettant une programmation graphique au lieu de textuels. Presque la moitié des articles (17) n'ont pas mentionné l'outil employé, ce qui peut être considéré comme une limite de cette étude.

D'après les résultats, les caractéristiques de l'I4.0 dans la PCP-AA pourraient être divisées en trois niveaux par rapport à leur usage : à un premier niveau, l'aOr s'impose comme la caractéristique la plus abordée (26 usages). Cela était prévisible, car elle est liée directement à la PCP. A un deuxième niveau, la gKn, l'aRe et l'aAp (14, 10 et 9 usages respectivement) se présentent comme des caractéristiques moyennement traitées. Finalement, un troisième niveau contient les cEn, les cCl et l'aEx (3, 2 et 1 usage respectivement). Ce niveau est marqué par une très rare utilisation dans la littérature. Les plus grandes perspectives de recherche se trouvent donc dans les caractéristiques appartenant au deuxième et troisième niveau. Cependant, il faut mentionner que le premier niveau propose aussi des perspectives de recherche. Cela peut être perçu dans la figure 6, où certains cas d'application ne sont pas couverts par l'aOr.

Il est intéressant de remarquer que les cEn se trouvent dans ce troisième niveau. En effet, malgré les problématiques environnementales actuelles, la PCP-AA semble ne pas traiter suffisamment cette caractéristique. Cela soulève des perspectives de recherche intéressantes en raison de la diversité des applications possibles et du besoin urgent de prendre en compte les enjeux environnementaux dans l'I4.0.

Finalement, la figure 6 montre que les domaines de la PCP-AA dans l'I4.0 sont couverts à la hauteur de 49 %. De ce fait, les perspectives de recherche de cette étude se concentreront sur les domaines (un cas d'application visant une caractéristique de l'I4.0 précise) qui n'ont pas été abordés ou qui ont été peu traités. En outre, des études théoriques seront menées afin de caractériser davantage d'éléments proposés par Zellner (2011) dans le concept des « Eléments Indispensables d'une Méthode ». Cela permettra de proposer une méthode pour implémenter une PCP-AA qui pourra servir de référence à d'autres travaux.

# 6 REFERENCES

Altaf, M. S., Bouferguene, A., Liu, H., Al-Hussein, M., Yu, H., (2018) Integrated production planning and control system for a panelized home prefabrication facility using simulation and RFID. *Automation in Construction*, Elsevier, 85 (February 2017), pp. 369–383.

Diaz-Rozo, J., Bielza, C., Larrañaga, P., (2017) Machine Learning-based CPS for Clustering High throughput Machining Cycle Conditions. *Procedia Manufacturing*, 10, pp. 997–1008.

Dolgui, A., Bakhtadze, N., Pyatetsky, V., Sabitov, R., Smirnova, G., Elpashev, D., Zakharov, E., (2018) Data Mining-Based Prediction of Manufacturing Situations Data Mining-Based. *IFAC-PapersOnLine*, Elsevier B.V., 51 (11), pp. 316–321.

Gyulai, D., Kádár, B., Monosotori, L., (2015) Robust production planning and capacity control for flexible assembly lines. *IFAC-PapersOnLine*, Elsevier Ltd., 28 (3), pp. 2312–2317.

Gyulai, D., Kádár, B., Monostori, L., (2014) Capacity planning and resource allocation in assembly systems consisting of dedicated and reconfigurable lines. *Procedia CIRP*, Elsevier B.V., 25 (C), pp. 185–191.

Gyulai, D., Pfeiffer, A., Nick, G., Gallina, V., Sihn, W., Monostori, L., (2018) Lead time prediction in a flow-shop environment with analytical and machine learning approaches. *IFAC-PapersOnLine*, 51 (11), pp. 1029–1034.

Hammami, Z., Mouelhi, W., Said, L. Ben, (2016) A self adaptive neural agent based decision support system for solving dynamic real time scheduling problems. *Proceedings - The 2015 10th International Conference on Intelligent Systems and Knowledge Engineering, ISKE 2015*, pp. 494–501.

Ji, W., Wang, L., (2017) Big data analytics based fault prediction for shop floor scheduling. *Journal of Manufacturing Systems,* The Society of Manufacturing Engineers, 43, pp. 187–194.

Jordan, M. I., Mitchell, T. M., (2015) Machine learning : Trends, perspectives, and prospects, 349 (6245).

Kartal, H., Oztekin, A., Gunasekaran, A., Cebi, F., (2016) An integrated decision analytic framework of machine learning with multi-criteria decision making for multi-attribute inventory classification. *Computers and Industrial Engineering,* Elsevier Ltd, 101, pp. 599–613.

Kho, D. D., Lee, S., Zhong, R. Y., (2018) Big Data Analytics for Processing Time Analysis in an IoT-enabled manufacturing Shop Floor. *Procedia Manufacturing,* Elsevier B.V., 26, pp. 1411–1420.

Kohler, D., Weisz, J.-D., (2016) Industrie 4.0 - Les défis de la transformation numérique du modèle industriel allemand [Industry 4.0: The Challenges of the Digital Transformation of the German Industrial Model], dans La Documentation française (ed.). Paris, p. 176.

Kruger, G. H., Shih, A. J., Hattingh, D. G., Van Niekerk, T. I., (2011) Intelligent machine agent architecture for adaptive control optimization of manufacturing processes. *Advanced Engineering Informatics*, Elsevier Ltd, 25 (4), pp. 783–796.

Kusiak, A., (2017) Smart manufacturing must embrace big data. *Nature*, 544 (7648), pp. 23–25.

Lai, L. K. C., Liu, J. N. K., (2012) WIPA : Neural network and case base reasoning models for allocating work in progress. *Journal of Intelligent Manufacturing*, 23 (3), pp. 409–421.

Leng, J., Chen, Q., Mao, N., Jiang, P., (2018) Combining granular computing technique with deep learning for service planning under social manufacturing contexts. *Knowledge-Based Systems*, Elsevier B.V., 143, pp. 295–306.

Li, D. C., Chen, C. C., Chen, W. C., Chang, C. J., (2012a) Employing dependent virtual samples to obtain more manufacturing information in pilot runs. *International Journal of Production Research*, 50 (23), pp. 6886–6903.

Li, X., Wang, J., Sawhney, R., (2012b) Reinforcement learning for joint pricing, lead-time and scheduling decisions in make-to-order systems. *European Journal of Operational Research,* Elsevier B.V., 221 (1), pp. 99–109.

Lingitz, L., Gallina, V., Ansari, F., Gyulai, D., Pfeiffer, A., Sihn, W., (2018) Lead time prediction using machine learning algorithms: A case study by a semiconductor manufacturer. *Procedia CIRP*, 72, pp. 1051–1056.

Lubosch, M., Kunath, M., Winkler, H., (2018) Industrial scheduling with Monte Carlo tree search and machine learning. *Procedia CIRP,* Elsevier B.V., 72, pp. 1283–1287.

Lv, Y., Qin, W., Yang, J., Zhang, J., (2018) Adjustment mode decision based on support vector data description and evidence theory for assembly lines. *Industrial Management and Data Systems*, 118 (8), pp. 1711–1726.

Manns, M., Wallis, R., Deuse, J., (2015) Automatic proposal of assembly work plans with a controlled natural language.

*Procedia CIRP*, 33, pp. 345–350.

Mitchell, T., (1997) Machine Learning. in. McGraw-Hill, p. 2.

Mori, J., Mahalec, V., (2015) Planning and scheduling of steel plates production. Part I : Estimation of production times via hybrid Bayesian networks for large domain of discrete variables. *Computers and Chemical Engineering*, Elsevier Ltd, 79, pp. 113–134.

Palombarini, J., Martínez, E., (2012) SmartGantt - An intelligent system for real time rescheduling based on relational reinforcement learning. *Expert Systems with Applications*, Elsevier Ltd, 39 (11), pp. 10251–10268.

Qu, S., Wang, J., Govil, S., Leckie, J. O., (2016) Optimized Adaptive Scheduling of a Manufacturing Process System with Multi-skill Workforce and Multiple Machine Types: An Ontology-based, Multi-agent Reinforcement Learning Approach. *Procedia CIRP*, Elsevier B.V., 57, pp. 55–60.

Reboiro-Jato, M., Glez-Dopazo, J., Glez, D., Laza, R., Gálvez, J. F., Pavón, R., Glez-Peña, D., Fdez-Riverola, F., (2011) Using inductive learning to assess compound feed production in cooperative poultry farms. *Expert Systems with Applications*, 38(11), pp. 14169–14177.

Reuter, C., Brambring, F., Weirich, J., Kleines, A., (2016) Improving Data Consistency in Production Control by Adaptation of Data Mining Algorithms. *Procedia CIRP*, 56, pp. 545–550.

Rostami, H., Blue, J., Yugma, C., (2018) Automatic equipment fault fingerprint extraction for the fault diagnostic on the batch process data. *Applied Soft Computing,* Elsevier B.V., 68, pp. 972–989.

Ruessmann, M., Lorenz, M., Gerbert, P., Waldner, M., Justus, J., Engel, P., Harnisch, M., (2015) Industry 4.0: The Future of Productivity and Growth in Manufacturing. *The Boston Consulting Group, Vol. 9*.

Schuh, G., Prote, J. P., Luckert, M., Hünnekes, P., (2017) Knowledge Discovery Approach for Automated Process Planning. *Procedia CIRP*, 63, pp. 539–544.

Shahzad, A., Mebarki, N., (2012) Data mining based job dispatching using hybrid simulation-optimization approach for shop scheduling problem. *Engineering Applications of Artificial Intelligence*, Elsevier, 25 (6), pp. 1173–1181.

Solti, A., Raffel, M., Romagnoli, G., Mendling, J., (2018) Misplaced product detection using sensor data without planograms. *Decision Support Systems*, 112 (June), pp. 76–87.

Stricker, N., Kuhnle, A., Sturm, R., Friess, S., (2018) Reinforcement learning for adaptive order dispatching in the semiconductor industry. *CIRP Annals*, CIRP, 67 (1), pp. 511–514.

Tao, F., Qi, Q., Liu, A., Kusiak, A., (2018) Data-driven smart manufacturing. *Journal of Manufacturing Systems*, 48, pp. 157–169.

Tian, G., Zhou, M., Chu, J., (2013) A chance constrained programming approach to determine the optimal disassembly sequence. *IEEE Transactions on Automation Science and Engineering*, 10 (4), pp. 1004–1013.

Tong, Y., Li, J., Li, S., Li, D., (2016) Research on Energy-Saving Production Scheduling Based on a Clustering Algorithm for a Forging Enterprise. *Sustainability*, 8 (2), p. Article number 136.

Tony Arnold, J. R., Chapman, S. N., Clive, L. M., (2012) Introduction to Materials Management. in. Pearson, p. 118.

Tranfield, D., Denyer, D., Smart, P., (2003) Towards a Methodology for Developing Evidence-Informed Management Knowledge by Means of Systematic Review. *British Journal of Management*, 14, pp. 207–222.

Tuncel, E., Zeid, A., Kamarthi, S., (2012) Solving large scale disassembly line balancing problem with uncertainty using reinforcement learning. *Journal of Intelligent Manufacturing*, 25 (4), pp. 647–659.

Wang, C. L., Rong, G., Weng, W., Feng, Y. P., (2015) Mining scheduling knowledge for job shop scheduling problem. *IFAC-PapersOnLine*, Elsevier Ltd., 28 (3), pp. 800–805.

Wang, J., Ma, Y., Zhang, L., Gao, R. X., Wu, D., (2018a) Deep learning for smart manufacturing: Methods and applications. *Journal of Manufacturing Systems,* The Society of Manufacturing Engineers, 48, pp. 144–156.

Wang, J., Yang, J., Zhang, J., Wang, X., Zhang, W., (2018b) Big data driven cycle time parallel prediction for production planning in wafer manufacturing. *Enterprise Information Systems,* Taylor & Francis, 12 (6), pp. 714–732.

Wang, J., Zhang, J., Wang, X., (2018c) Bilateral LSTM : A two-dimensional long short-term memory model with multiply memory units for short-term cycle time forecasting in re-entrant manufacturing systems. *IEEE Transactions on Industrial Informatics*, 14 (2), pp. 748–758.

Waschneck, B., Bauernhansl, T., Knapp, A., Kyek, A., (2018) Optimization of global production scheduling with deep reinforcement learning, 72, pp. 1264–1269.

Wauters, T., Verbeeck, K., Verstraete, P., Vanden Berghe, G., De Causmaecker, P., (2012) Real-world production scheduling for the food industry: An integrated approach. *Engineering Applications of Artificial Intelligence*, Elsevier, 25 (2), pp. 222–228.

Yang, Z., Zhang, P., Chen, L., (2016) RFID-enabled indoor positioning method for a real-time manufacturing execution system using OS-ELM. *Neurocomputing*, Elsevier, 174, pp. 121–133.

Zellner, G., (2011) A structured evaluation of business process improvement approaches. *Business Process Management Journal*, 17 (2), pp. 203–237.

Zhang, Z., Zheng, L., Hou, F., Li, N., (2011) Semiconductor final test scheduling with Sarsa (λ, k) algorithm. *European Journal of Operational Research*, 215 (2), pp. 446–458.

Zhong, R. Y., Huang, G. Q., Dai, Q., (2013) Mining standard operation times for real-time advanced production planning and scheduling from RFID-enabled shopfloor data. *IFAC Proceedings Volumes (IFAC-PapersOnline)*.